\documentclass[letterpaper, 10 pt, conference]{ieeeconf}  

\IEEEoverridecommandlockouts                              

\usepackage{graphicx}            
\usepackage{booktabs}            
\usepackage[table]{xcolor}       
\usepackage{pifont}              
\usepackage[normalem]{ulem}      
\usepackage{afterpage}           
\usepackage{fontawesome5}        
\makeatletter
\let\NAT@parse\undefined
\makeatother
\usepackage[colorlinks=true,linkcolor=black,urlcolor=black]{hyperref}

\definecolor{foreactblue}{RGB}{56,101,172}
\definecolor{foreactpurple}{RGB}{116,60,145}
\hypersetup{citecolor=foreactpurple}

\newif\ifanonymoussubmission
\anonymoussubmissionfalse

\newcommand{\cmark}{\ding{51}}
\newcommand{\xmark}{\ding{55}}

\newcommand{\best}[1]{\textbf{#1}}
\newcommand{\second}[1]{\uline{#1}}
\newcommand{\third}[1]{\dashuline{#1}}

\definecolor{groupblue}{RGB}{235,242,249}
\definecolor{ourscolor}{RGB}{241,238,249}

\makeatletter

\newcommand{\captionof}[1]{\def\@captype{#1}\caption}

\let\originalmakecaption\@makecaption

\long\def\@makecaption#1#2{%
  \ifx\@captype\@IEEEtablestring
    \parbox[t]{\hsize}{%
      \footnotesize
      \noindent #1: #2\par
    }%
    \vspace{3pt}%
  \else
    \originalmakecaption{#1}{#2}%
  \fi
}

\makeatother

\title{\texorpdfstring{%
\resizebox{\textwidth}{!}{\bfseries\textcolor{foreactblue}{Fore}\textcolor{foreactpurple}{Act}: Geometry-Guided Action Learning from World Representations}%
}{ForeAct: Geometry-Guided Action Learning from World Representations}}

\ifanonymoussubmission
\author{Anonymous Authors}
\else
\author{%
\authorblockN{Wei Lu$^{1,*}$, Jiaxi Zhang$^{1,*}$, Yunheng Wang$^{1,*}$,
Taowen Wang$^{1}$, Yuetong Fang$^{1}$, Zhenhui Zhang$^{1}$\\
Shuning Zhang$^{1}$, Zizhao Yuan$^{1}$, Yixiao Feng$^{1}$,
Zecui Zeng$^{2,\dagger}$, Renjing Xu$^{1,\dagger}$}
\authorblockA{$^{1}$The Hong Kong University of Science and Technology (Guangzhou)\\
$^{2}$JD Explore Academy\\[0.25em]
\href{https://github.com/Bulrush666/ForeAct}{\faGithub\enspace Github}\qquad
\href{https://huggingface.co/Bulru1234/ForeAct}{\raisebox{-0.15em}{\includegraphics[height=1.05em]{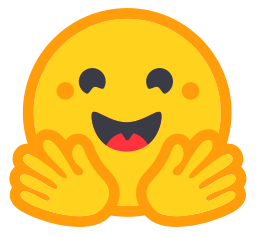}}\enspace HuggingFace}}
\thanks{$^{*}$Equal contribution. $^{\dagger}$Co-corresponding authors.}%
}
\fi

\begin{document}

\maketitle
\thispagestyle{empty}
\pagestyle{empty}

\begingroup
\color{black}
\begin{abstract}
Robotic manipulation requires representations that capture both scene geometry and its evolution during interaction. Existing approaches have largely treated these two aspects separately. World action models capture such dynamics by coupling action prediction with future video generation, but their reliance on iterative diffusion or rectified-flow inference leads to high computational cost, while their representations lack explicit spatial structure. Explicitly modeling scene geometry typically requires auxiliary modules that increase both parameter count and computational cost. Conversely, geometry-centric policies primarily recover observed spatial structure and often lack explicit supervision for anticipating future geometric changes during interaction. To address these limitations, we introduce \textcolor{foreactblue}{Fore}\textcolor{foreactpurple}{Act}, which repurposes a pretrained video world model as a single-pass representation backbone for low-latency control. \textcolor{foreactblue}{Fore}\textcolor{foreactpurple}{Act} directly processes the clean latent of the current observation in one forward pass, eliminating future-frame synthesis and iterative denoising while retaining the backbone's pretrained spatiotemporal priors. Rather than introducing a separate geometry backbone, \textcolor{foreactblue}{Fore}\textcolor{foreactpurple}{Act} uses learnable tokens supervised by next-frame relative depth to shape the shared video representation through bidirectional self-attention. This compact interface turns the spatial knowledge already encoded in the backbone into a future-aware representation for action generation, with little parameter overhead. The lightweight geometry readout is used only during training and removed at deployment, enabling low-latency, geometry-aware control without depth input or an external geometry backbone. Without additional large-scale pretraining, \textcolor{foreactblue}{Fore}\textcolor{foreactpurple}{Act} achieves 96.6\% average success on LIBERO, 63.5\% on SimplerEnv-Bridge, and a 74.4\% success rate across 90 real-world rollouts.
\end{abstract}
\endgroup

\begingroup
\color{red}
{\color{black}\section{INTRODUCTION}}

\begin{figure}[t]
    \centering
    \includegraphics[width=\columnwidth]{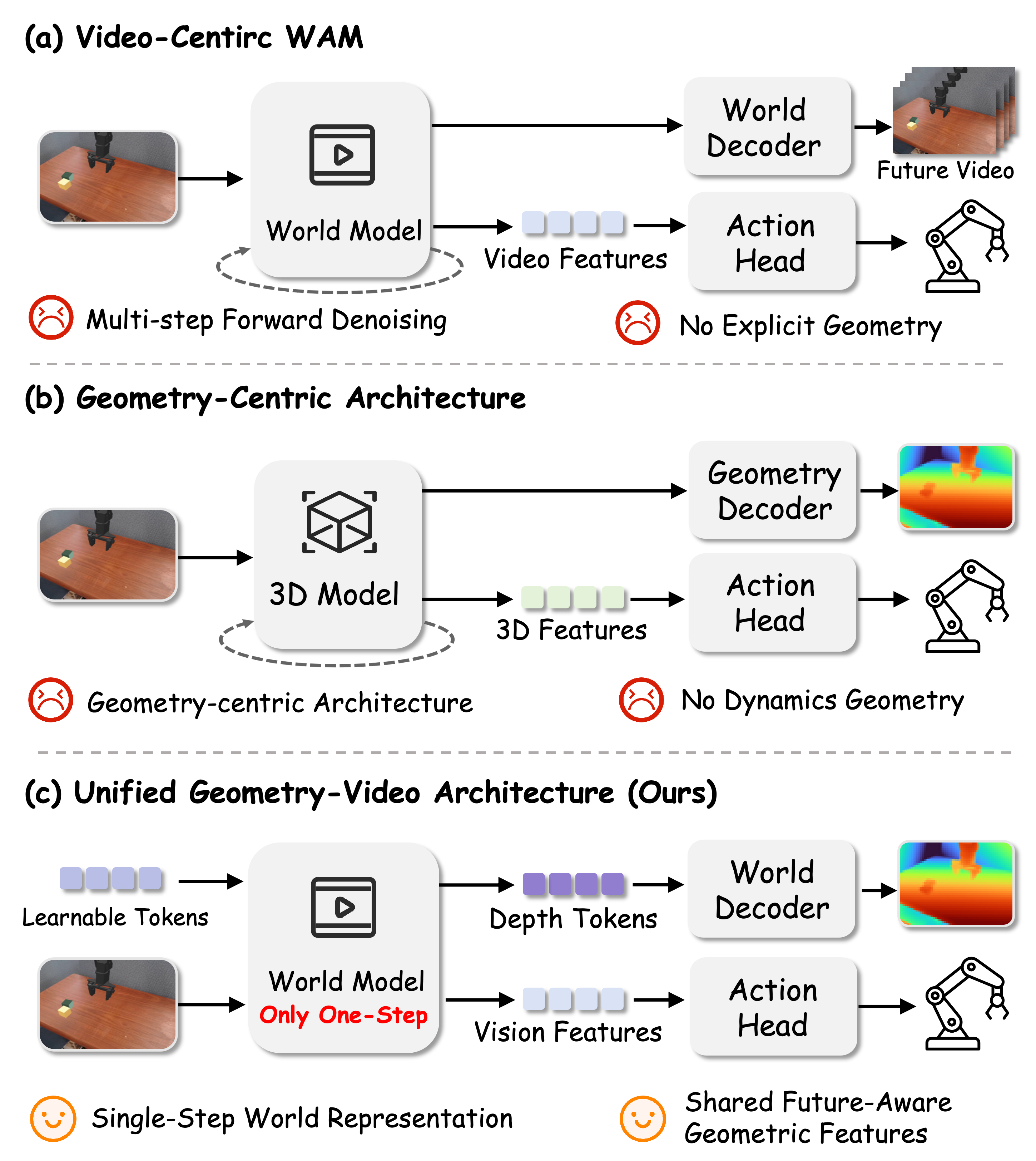}
    \caption{
    \textcolor{black}{Backbone paradigms for robot control. (a) Video-centric WAMs derive action features through iterative denoising. (b) Geometry-centric policies use features from pretrained geometric backbones. (c) ForeAct extracts action features in a single video-backbone pass, shaped by next-frame depth supervision during training.}
    }
    \label{fig:intro3}
\end{figure}



\color{black}

Pretrained foundation models have become an important basis for robotic manipulation, enabling policies to transfer knowledge acquired from large-scale visual data to action learning. Vision-language-action models demonstrate the value of this approach by grounding semantic and visual representations in robot behavior~\cite{brohan2023rt2,kim2024openvla,black2024pi0}. Beyond semantic understanding, effective manipulation requires representations that capture both scene geometry and its evolution during interaction. Video generative models offer temporal priors for anticipating scene changes, while geometric foundation models provide spatial representations relevant to precise interaction. These capabilities motivate a unified policy representation that captures spatial structure and its evolution during physical interaction while remaining efficient enough for closed-loop control.

\color{red}


\color{black}
Existing approaches incorporate these capabilities through different backbone architectures. Video-centric world action models (WAMs; Fig.~\ref{fig:intro3}(a)) adapt pretrained video generative models by coupling action prediction with future visual generation. Through iterative diffusion or rectified-flow inference, they obtain representations of anticipated visual states that transfer motion and temporal priors to control~\cite{ye2026dreamzero,kim2026cosmospolicy,yuan2026fastwam}. Geometry-centric policies (Fig.~\ref{fig:intro3}(b)) instead extract features from pretrained geometric backbones, using their encoded scene structure and spatial relationships to guide action prediction~\cite{abouzeid2025geoawarevla,ge2025vggtdp,song2026vga}. These representations provide strong spatial priors for precise manipulation.

Despite these advances, two limitations remain.
\textbf{\ding{182}} \textbf{Costly Inference and Limited Spatial Structure.} Video-centric WAMs capture interaction dynamics by coupling action prediction with future video generation. However, their reliance on iterative diffusion or rectified-flow inference incurs high computational cost, while their representations lack explicit spatial structure.
\textbf{\ding{183}} \textbf{Limitations of Explicit Geometric Modeling.} Explicitly modeling scene geometry typically requires auxiliary modules, increasing both parameter count and computational cost. Geometry-centric policies instead recover observed spatial structure through pretrained geometric backbones, but have limited capacity to capture action-conditioned spatiotemporal dynamics. Together, these limitations call for a unified representation that captures scene geometry and interaction dynamics without the computational burden of iterative generation or an additional geometric backbone.


\color{red}
{\color{black}

Motivated by evidence that visual generative features transfer to geometric tasks~\cite{ke2024marigold,xu2024genpercept} and that video-model representations support geometric estimation~\cite{wang2026genception}, we seek to expose and adapt the spatial knowledge within a pretrained video backbone alongside its temporal priors. We introduce ForeAct (Fig.~\ref{fig:intro3}(c)), which repurposes a pretrained video world model as a unified representation backbone through two complementary designs. \textbf{\ding{182}} \textbf{Single-Step World Representation.} ForeAct processes the clean observation latent at $t_{\mathrm{diff}}=0$ in one video-backbone forward pass, obtaining observation-anchored action features that draw on pretrained spatiotemporal priors without future-frame synthesis or iterative video denoising. \textbf{\ding{183}} \textbf{Shared Future-Aware Geometric Features.} A compact set of learnable geometry tokens interacts with visual tokens through bidirectional self-attention and receives next-frame relative-depth supervision. This interaction conveys predictive geometric supervision to the shared action features, adapting the backbone's spatial knowledge for future-aware control without a separate geometric encoder. The lightweight geometry readout is used only during training and removed at deployment, while geometry tokens remain within the shared backbone, enabling efficient geometry-aware control without depth input. Our contributions are:

}
\color{red}
\color{black}


\begin{itemize}
    \item We propose ForeAct, which repurposes a pretrained video world model as a single-pass representation backbone for robotic manipulation, extracting action features for low-latency control without future-frame synthesis or iterative video denoising.

    \item We introduce a compact geometry-token interface that shapes shared action representations through next-frame relative-depth supervision and bidirectional self-attention, without requiring depth input or an external geometric backbone at deployment.

    \item We evaluate ForeAct on LIBERO, SimplerEnv-Bridge, and six real-world manipulation tasks, achieving success rates of $96.6\%$, $63.5\%$, and $74.4\%$, respectively, without additional large-scale pretraining. Controlled ablations validate its core components and design choices.
\end{itemize}

\color{red}

\begin{figure*}[t]
    \centering
    \includegraphics[width=0.98\textwidth]{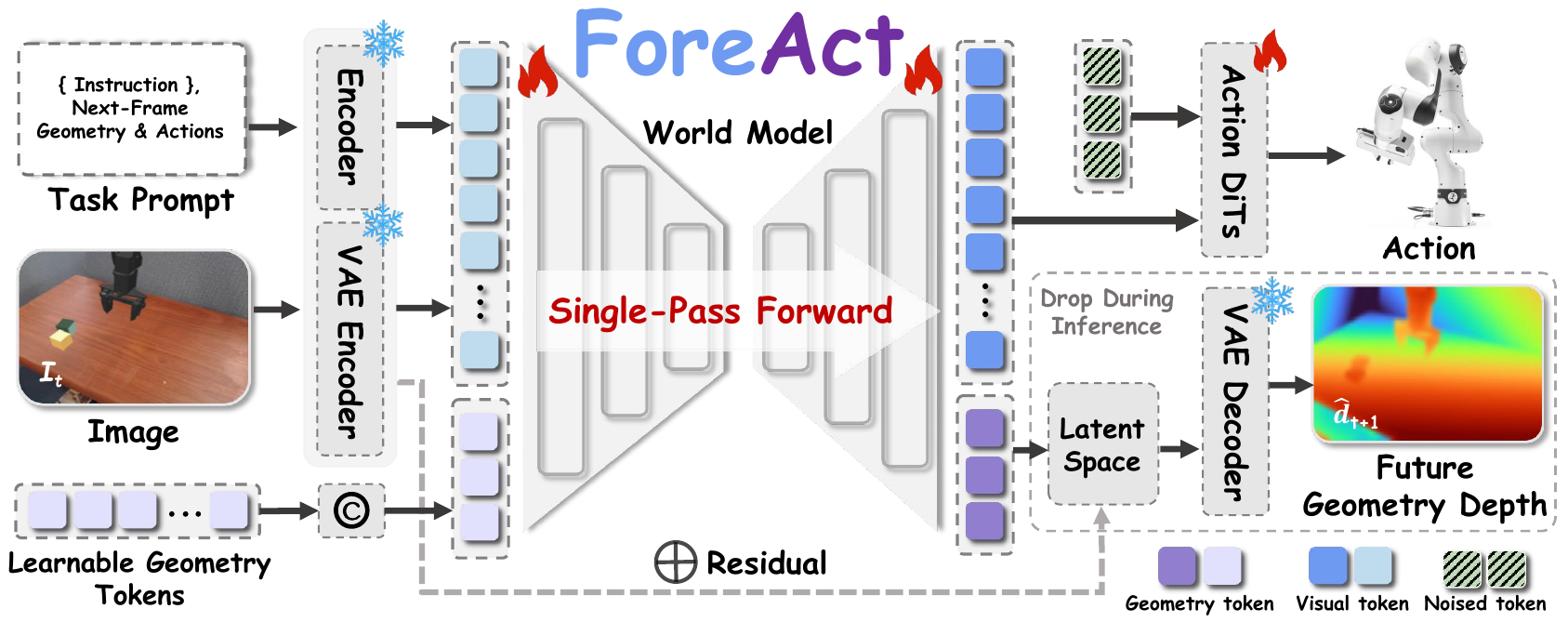}
    \caption{
        \textcolor{black}{
            \textbf{ForeAct architecture.}
            A single forward pass through the pretrained world model jointly
            processes the clean visual latent, language embedding, and learnable
            future-geometry tokens. The output visual tokens are passed to the
            Action DiT, while the geometry tokens are projected to a spatial
            latent and decoded into next-frame relative depth
            $\hat{d}_{t+1}$ for training supervision. The geometry readout is
            removed at inference, preserving efficient single-pass control
            without depth decoding.
        }
    }
    \label{fig:pipeline}
\end{figure*}

\endgroup

\begingroup
\color{red}
{\color{black}\section{RELATED WORK}}

\color{black}
\subsection{Vision-Language-Action Models}

Vision-language-action (VLA) models align the semantic representations of large-scale vision-language
pre-training with robot action spaces, establishing a common interface from visual observations and
language instructions to control. Early approaches discretize actions as language tokens
~\cite{brohan2023rt2,zhang2024grape}, whereas later architectures predict continuous action chunks and
scale training across heterogeneous robot data and embodiments
~\cite{ghosh2024octo,kim2024openvla,li2024cogact,black2024pi0,shukor2025smolvla}. This progression
improves semantic grounding, instruction following, and transfer across tasks and embodiments
~\cite{bjorck2025groot}. However, most VLA policies
still condition the action decoder primarily on the current observation and language, leaving the temporal
consequences of interaction and the geometry of future scene configurations to be learned implicitly from
robot demonstrations. ForeAct preserves the general VLA action interface while providing the action decoder with a predictive, geometry-aware representation extracted from a pretrained video world model, thereby introducing explicit spatiotemporal structure into action learning.

\color{black}
\subsection{World Action Models}

World Action Models (WAMs) extend the reactive VLA interface by jointly modeling future observations and robot actions. Their predicted future states provide temporal context for planning and long-horizon control. Representative systems jointly predict future images and
robot actions, use visual goals as intermediate reasoning states, or couple action learning to conditional
video generation~\cite{wu2023gr1,cen2025worldvla,zhao2025cotvla,ye2026dreamzero,kim2026cosmospolicy}.
Fast-WAM~\cite{yuan2026fastwam} further suggests that the control benefit can arise from the learned world representation rather
than from the final sampled frames. A central limitation of diffusion-based WAMs,
however, is that action decoding is tied to noisy latents or intermediate denoising states indexed by the
noise realization and diffusion timestep. These synthesis-oriented features yield a stochastic control
state and require iterative future-frame sampling~\cite{ye2026dreamzero}. ForeAct processes the clean current observation in a single world-model forward pass and passes the resulting visual tokens to the Action DiT, avoiding noisy future generation while enabling efficient control.

\color{black}
{\color{black}\subsection{Geometry-Aware Policy Learning}}


Explicit 3-D representations provide spatial relations essential for contact-rich manipulation and precise
placement. GeoAware-VLA~\cite{abouzeid2025geoawarevla}, VGGT-DP~\cite{ge2025vggtdp}, and VGA~\cite{song2026vga} transfer representations from pretrained geometric models
to action prediction, establishing geometric
perception as a backbone paradigm for policy learning. GP3~\cite{qian2025gp3} likewise explores geometry for robotic
control. Future geometry is also explicitly modeled by predictive approaches such as
GAM~\cite{han2026gam} and PointWorld~\cite{huang2026pointworld}. Our distinction is therefore the pretrained
backbone and how its priors are adapted, rather than the presence of geometric prediction alone.
ForeAct starts from video generative pretraining and makes the backbone's spatial knowledge predictive
through next-frame depth supervision. It introduces learnable future-geometry tokens into this video world model and
supervises them with next-frame relative depth. Full self-attention allows the depth-supervised geometry tokens to shape the visual features used for action decoding, integrating spatial reasoning and control without an external geometric encoder or depth input at deployment.

\endgroup

\section{Method}
\label{sec:method}

In this section, we first present the overall design of ForeAct (Sec.~III-A). As illustrated in Fig.~\ref{fig:pipeline}, ForeAct takes an RGB observation and a language instruction as inputs and couples a pretrained video world model with action generation and predictive geometric supervision. We then detail the two central components of this framework: Latent World Representation (Sec.~III-B), which extracts an observation-anchored representation from a single clean world model forward pass, and Future Geometry Learning (Sec.~III-C), which introduces geometry tokens and next-frame depth supervision to shape the shared representation for control.

\subsection{ForeAct Overview}

\textbf{Problem Formulation.} Given an RGB observation $\mathbf{I}_t$ and a language instruction $c$, ForeAct predicts an $H$-step action chunk $\mathbf{a}_{t:t+H-1}$ without requiring proprioceptive state and, during training, jointly estimates the relative depth of the next observation. The task prompt augments $c$ with ``Next-Frame Geometry \& Actions'' (Fig.~\ref{fig:pipeline}). We formulate this joint prediction as follows:
\begin{equation}
    (\hat{\mathbf{a}}_{t:t+H-1},\hat{\mathbf{d}}_{t+1})
    =\mathcal{F}_{\theta}(\mathbf{I}_t,c),
    \label{eq:overview_problem}
\end{equation}
where $H$ is the action horizon and $\hat{\mathbf{d}}_{t+1}$ denotes the normalized relative depth of the immediate next observation. This state-free formulation supports deployment when robot-state measurements are unavailable or vary across embodiments, while the auxiliary depth target encourages the learned representation to capture short-horizon scene evolution. Depth supervision is used only during training and is never provided to the action policy at inference.

\textbf{Model Architecture.} As illustrated in Fig.~\ref{fig:pipeline}, ForeAct employs a shared latent world backbone $\mathcal{W}{\theta}$ whose representations are processed by an action expert $\mathcal{A}{\theta}$ and a future-geometry head $\mathcal{G}_{\theta}$. The shared backbone maps the current observation and language instruction to a latent world representation shared by both decoders:
\begin{equation}
    (\hat{\mathbf{a}}_{t:t+H-1},\hat{\mathbf{d}}_{t+1})
    =\bigl[(\mathcal{A}_{\theta},\mathcal{G}_{\theta})
    \circ\mathcal{W}_{\theta}\bigr](\mathbf{I}_t,c).
    \label{eq:overview_architecture}
\end{equation}
The action expert maps the shared latent representation to executable
controls, while the geometry head predicts the relative depth of the
next observation during training. This auxiliary objective directly
regularizes the shared representation used for action prediction.
The geometry head is used exclusively for training, whereas inference relies only on the world backbone and action expert. These two components are described in Secs.~III-B and III-C, respectively.

\textbf{Training Objective.} ForeAct jointly optimizes action generation and predictive geometry. For each ground-truth action chunk, conditional flow matching constructs an interpolant between Gaussian noise and the normalized action target, and the action expert regresses the corresponding transport velocity. The geometry branch uses the cached normalized relative depth of the next observation, encoded by the frozen VAE, and is supervised in latent, decoded pixel, and spatial-gradient spaces. The two branches are combined as follows:
\begin{equation}
    \mathcal{L}=\mathcal{L}_{\mathrm{act}}+\lambda_{\mathrm{geo}}
    \left(\lambda_{\mathrm{lat}}\mathcal{L}_{\mathrm{lat}}+\lambda_{\mathrm{pix}}\mathcal{L}_{\mathrm{pix}}+\lambda_{\mathrm{grad}}\mathcal{L}_{\mathrm{grad}}\right),
    \label{eq:overview_training_objective}
\end{equation}
with $\lambda_{\mathrm{geo}}=0.01$, $\lambda_{\mathrm{lat}}=1$, $\lambda_{\mathrm{pix}}=0.1$, and $\lambda_{\mathrm{grad}}=0.05$. Depth is auxiliary supervision only, and its gradient to the shared Cosmos backbone is scaled by $0.01$.

\begin{figure}[t]
    \centering
    \includegraphics[width=\columnwidth]{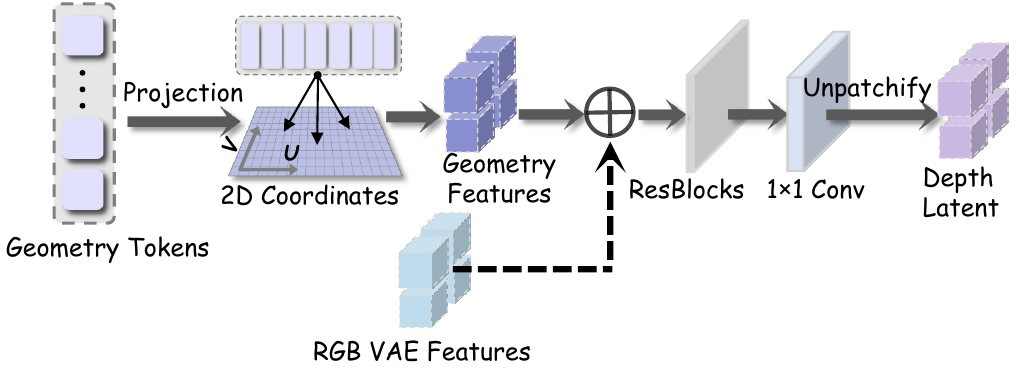}
    \caption{
    \textbf{Future-Geometry Latent Readout.} Geometry tokens are projected into a global context and combined with 2D coordinates to form spatial features. Fused with VAE-derived residual features from the current observation, they are refined and mapped to a next-frame depth latent.
    }
    \label{fig:method1}
\end{figure}

\subsection{Latent World Representation}
Diffusion-based world action models (WAMs) commonly obtain predictive context by denoising noise-initialized future latents and coupling intermediate generative features to action prediction~\cite{ye2026dreamzero,kim2026cosmospolicy,yuan2026fastwam}. While effective for visual forecasting, this formulation makes the control representation depend on sampled noise and diffusion timesteps, incurs repeated backbone evaluations, and prioritizes future synthesis over observation-aligned evidence. Recent findings further suggest that the control benefit of video modeling arises largely from its learned representation rather than the final generated frames~\cite{yuan2026fastwam,wang2026genception}. ForeAct therefore uses the pre-trained video world model as a conditional representation backbone: a clean observation latent is processed once at the fixed diffusion endpoint to directly expose its spatiotemporal prior. The resulting representation is stable, anchored to the current scene, and efficient for action conditioning without explicit future-frame sampling.

As illustrated in Fig.~\ref{fig:pipeline}, the current RGB observation $\mathbf{I}_t$ is encoded by the frozen video VAE into a clean visual latent $\mathbf{z}_t$, while the frozen text encoder maps the language instruction $c$ to $\mathbf{e}_c$. After patch embedding, the visual tokens are concatenated with a set of learnable future-geometry tokens $\mathbf{Q}^{0}$ and processed once by the pre-trained world model backbone at the clean endpoint:
\begin{equation}
    [\mathbf{V}_t,\mathbf{Q}_t]
    = \mathcal{W}_{\theta}\!\left(
    [\mathcal{P}(\mathbf{z}_t);\mathbf{Q}^{0}],
    \mathbf{e}_c;\,t_{\mathrm{diff}}=0
    \right),
\label{eq:latent_world_representation}
\end{equation}
where $\mathbf{V}_t$ and $\mathbf{Q}_t$ denote the output visual and future-geometry tokens, respectively. The visual tokens $\mathbf{V}_t$ provide the observation-aligned world representation used to condition the flow-matching action DiT, whereas $\mathbf{Q}_t$ summarizes global spatial structure for the latent geometry branch. During training, the latter is combined with observation-derived VAE features and decoded into the relative depth of the next frame, allowing future geometric supervision to directly shape the shared world model representation. At inference, the geometry decoder is removed and actions are produced from the single-pass visual representation, requiring neither future-frame generation nor depth observations.


\providecommand{\best}[1]{\textbf{#1}}
\providecommand{\second}[1]{\uline{#1}}
\providecommand{\third}[1]{\dotuline{#1}}
\providecolor{groupblue}{RGB}{232,240,248}
\providecolor{ourscolor}{RGB}{242,239,247}

\newcommand{\renderliberocomparisontable}{%
\begingroup
\centering
\captionof{table}{
\textbf{Comparison on the LIBERO benchmark.}
Pre-Train denotes additional large-scale robot-action pre-training;
generic vision, language, video, or depth pre-training is excluded.
The best, second-best, and third-best results are highlighted.
}
\label{tab:libero_comparison}

\setlength{\tabcolsep}{2.6pt}
\renewcommand{\arraystretch}{1.03}

\resizebox{\linewidth}{!}{
\begin{tabular}{lcccccc}
\toprule
\textbf{Method}
& \textbf{Pre-Train}
& \textbf{Spatial}
& \textbf{Object}
& \textbf{Goal}
& \textbf{Long}
& \textbf{Avg.} \\
\midrule

\rowcolor{groupblue}
\multicolumn{7}{l}{\textit{Original VLAs}} \\
OpenVLA~\cite{kim2024openvla}
& \cmark & 84.7 & 88.4 & 79.2 & 53.7 & 76.5 \\
GRAPE~\cite{zhang2024grape}
& \cmark & 88.5 & 92.1 & 83.1 & 57.2 & 80.2 \\
ThinkAct~\cite{huang2025thinkact}
& \cmark & 88.3 & 91.4 & 87.1 & 70.9 & 84.4 \\
$\pi_0$-FAST~\cite{pertsch2025fast}
& \cmark & 96.4 & 96.8 & 88.6 & 60.2 & 85.5 \\
SmolVLA~\cite{shukor2025smolvla}
& \xmark & 93.0 & 94.0 & 91.0 & 77.0 & 88.8 \\
GR00T N1~\cite{bjorck2025groot}
& \cmark & 94.4 & 97.6 & 93.0 & 90.6 & 93.9 \\
GR00T N1.5~\cite{bjorck2025groot}
& \cmark & 91.4 & 97.6 & 94.0 & 85.6 & 92.1 \\
$\pi_0$~\cite{black2024pi0}
& \cmark & 96.8 & \second{98.8} & 95.8 & 85.2 & 94.2 \\
\midrule
\rowcolor{groupblue}
\multicolumn{7}{l}{\textit{Geometry-Aware VLAs}} \\
SpatialVLA~\cite{qu2025spatialvla}
& \cmark & 88.2 & 89.9 & 78.6 & 55.5 & 78.1 \\
MolmoAct-7B-D~\cite{lee2025molmoact}
& \cmark & 87.0 & 95.4 & 87.6 & 77.2 & 86.6 \\
4D-VLA~\cite{zhang2025fourdvla}
& \cmark & 88.9 & 95.2 & 90.9 & 79.1 & 88.6 \\
DepthVLA~\cite{yuan2025depthvla}
& \xmark & 96.4 & 98.0 & 95.8 & 89.2 & 94.9 \\
GeoPredict~\cite{qian2025geopredict}
& -- & 98.0 & \third{98.2} & 95.7 & 94.0 & 96.5 \\

\midrule
\rowcolor{groupblue}
\multicolumn{7}{l}{\textit{World Action Models}} \\
WorldVLA~\cite{cen2025worldvla}
& \xmark & 87.6 & 96.2 & 83.4 & 60.0 & 81.8 \\
FlowVLA~\cite{zhong2025flowvla}
& \xmark & 93.2 & 95.0 & 91.6 & 72.6 & 88.1 \\
DreamVLA~\cite{dreamvla25}
& \cmark & 97.5 & 94.0 & 89.5 & 89.5 & 92.6 \\
F1~\cite{lv2025f1}
& \cmark & \second{98.2} & 97.8 & 95.4 & 91.3 & 95.7 \\
GE-Act~\cite{liao2025genie}
& \cmark & \second{98.2} & 97.6 & 95.8 & \third{94.4} & 96.5 \\
Fast-WAM~\cite{yuan2026fastwam}
& \xmark & \second{98.2} & \best{100.0} & \third{97.0} & \second{95.2} & \second{97.6} \\
Cosmos Policy~\cite{kim2026cosmospolicy}
& \xmark & \third{98.1} & \best{100.0} & \best{98.2} & \best{97.6} & \best{98.5} \\

\midrule
\rowcolor{ourscolor}
\textbf{ForeAct (Ours)}
& \xmark
& \best{98.4}
& \second{98.8}
& \second{97.2}
& 91.8
& \third{96.6} \\

\bottomrule
\end{tabular}
}
\endgroup
}

\newcommand{\rendersimplerenvcomparisontable}{%
\begingroup
\centering

\captionof{table}{
\textbf{Performance Comparison on SimplerEnv (WidowX/Bridge).}
We evaluate models across four tasks: Spoon on Towel (\textbf{ST}),
Carrot on Plate (\textbf{CP}), Stack Blocks (\textbf{SB}), and
Eggplant in Basket (\textbf{EB}).
}
\label{tab:simplerenv_comparison}

\setlength{\tabcolsep}{3pt}
\renewcommand{\arraystretch}{1.05}

\resizebox{\linewidth}{!}{
\begin{tabular}{lcccccc}
\toprule

\textbf{Models}
& \textbf{Pre-Train}
& \textbf{ST}
& \textbf{CP}
& \textbf{SB}
& \textbf{EB}
& \textbf{Avg.} \\

\midrule

Magma~\cite{yang2025magma}
& \cmark & 37.5 & 31.0 & 12.7 & 60.5 & 35.4 \\

RT-1-X~\cite{brohan2023rt1x}
& \cmark & 0.0 & 4.2 & 0.0 & 0.0 & 1.1 \\

Octo-Base~\cite{ghosh2024octo}
& \cmark & 15.8 & 12.5 & 0.0 & 41.7 & 17.5 \\

Octo-Small~\cite{ghosh2024octo}
& \cmark & 41.7 & 8.2 & 0.0 & 56.7 & 26.7 \\

OpenVLA~\cite{kim2024openvla}
& \cmark & 4.2 & 0.0 & 0.0 & 12.5 & 4.2 \\

ThinkAct~\cite{huang2025thinkact}
& \cmark & 58.3 & 37.5 & 8.7 & 70.8 & 43.8 \\

CogACT~\cite{li2024cogact}
& \cmark
& \third{71.7}
& \third{50.8}
& 15.0
& \second{67.5}
& \third{51.3} \\

SpatialVLA~\cite{qu2025spatialvla}
& \cmark
& 16.7
& 25.0
& \second{29.2}
& \best{100.0}
& 42.7 \\

$\pi_0$~\cite{black2024pi0}
& \cmark & 29.1 & 0.0 & \third{16.6} & 62.5 & 27.1 \\

$\pi_0$-FAST~\cite{pertsch2025fast}
& \cmark & 29.1 & 21.9 & 10.8 & \third{66.6} & 32.1 \\

GR00T N1.5~\cite{bjorck2025groot}
& \cmark
& \second{75.3}
& \best{54.3}
& \best{57.0}
& 61.3
& \second{61.9} \\

\midrule

\rowcolor{ourscolor}
\textbf{ForeAct (Ours)}
& \xmark
& \best{87.5}
& \second{54.2}
& 12.5
& \best{100.0}
& \best{63.5} \\

\bottomrule
\end{tabular}
}

\endgroup
}

\begin{figure*}[!t]
    \centering
    \includegraphics[width=\textwidth]{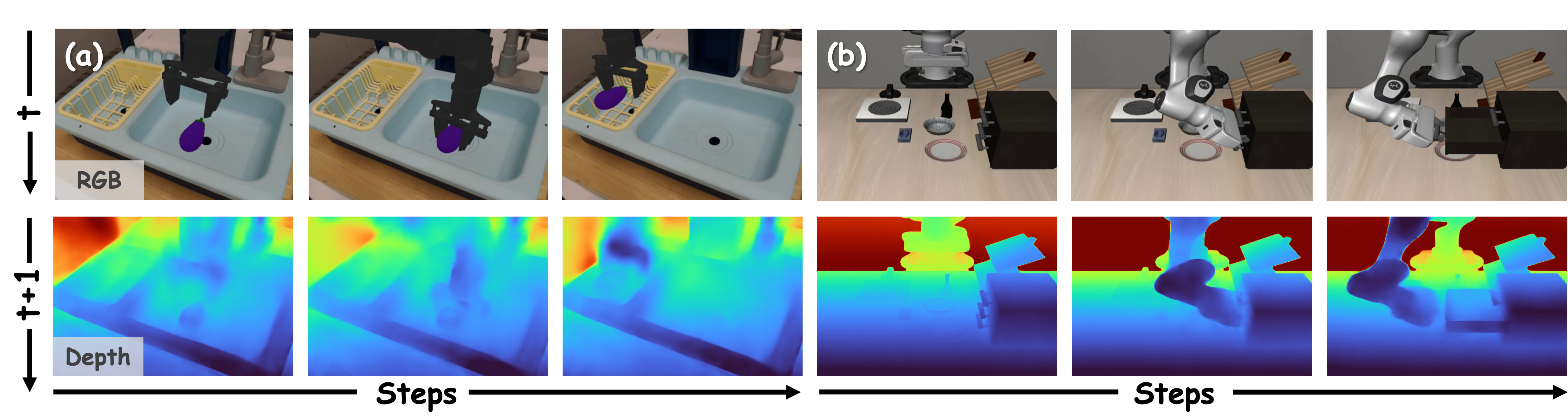}
    \caption{\textbf{Next-frame depth visualization on
    SimplerEnv--WidowX and LIBERO.}
    (a) SimplerEnv--WidowX. (b) LIBERO.
    Each benchmark shows early, middle, and late stages of
    one manipulation episode from left to right.
    The RGB row contains the current input frames $I_t$,
    while the Depth row shows ForeAct's predicted next-frame
    relative depth $\hat{D}_{t+1}$.
    The changing depth maps reflect the evolving
    robot--object configuration in both tasks.}
    \label{fig:depth_visualization}
\end{figure*}

\subsection{Future Geometry Learning}

Geometry-centric policies condition action prediction on spatial
representations transferred from geometric foundation
models~\cite{abouzeid2025geoawarevla,ge2025vggtdp,song2026vga}.
Such pretraining provides strong structural priors, while anticipating
future spatial changes requires predictive learning beyond reconstructing
observed geometry. ForeAct starts from the temporal priors of a pretrained
video world model and exposes its geometric knowledge through a compact
set of learnable future-geometry tokens $\mathbf{Q}^{0}$. Jointly updated
with visual tokens by unmasked self-attention, these tokens aggregate
spatial context and transmit next-frame depth supervision to the action
features. This aligns future geometry with action learning within the
video backbone, without requiring depth input or an additional geometric
encoder at inference.

As illustrated in Fig.~\ref{fig:method1}, the output geometry tokens
$\mathbf{Q}_t$ are first projected and concatenated into a global geometry
code $\mathbf{g}_t$, which is paired with the encoding $\gamma(u,v)$ of
each location on a low-resolution 2D coordinate grid $\Omega_s$. A
coordinate-conditioned mapper $\phi_{\mathrm{geo}}$ then instantiates
this global code over the grid to produce spatial geometry features
$\mathbf{F}_t^{\mathrm{geo}}$. Because this token-derived field captures
global structure but may omit observation-aligned details, the current
RGB VAE latent $\mathbf{z}_t$ is projected to the same resolution and
injected through a zero-initialized $1\!\times\!1$ residual mapping
$\mathcal{C}_0$. The complete latent-space decoder is summarized by
\begin{equation}
\hat{\mathbf{z}}_{t+1}^{d}
=
\mathcal{U}_{\phi}\!\left(
\mathbf{F}_t^{\mathrm{geo}}
+
\mathcal{C}_0\!\left(\mathcal{P}_z(\mathbf{z}_t)\right)
\right),
\label{eq:future_geometry_decoding}
\end{equation}
where $\mathcal{U}_{\phi}$ comprises the residual blocks,
$1\!\times\!1$ convolution, and unpatchify operation shown in
Fig.~\ref{fig:method1}. To construct the supervision target, the
normalized next-frame relative-depth map is replicated across three
channels and encoded by the frozen VAE encoder to obtain
$\mathbf{z}_{t+1}^{d}$ in the pretrained latent space. The geometry
readout is trained to align $\hat{\mathbf{z}}_{t+1}^{d}$ with
$\mathbf{z}_{t+1}^{d}$, allowing the paired frozen VAE decoder to
reconstruct the corresponding relative-depth map. The VAE thus serves
as a fixed latent-space codec rather than a learned depth estimator.
This supervision provides the geometry tokens with prospective
semantics, while the zero-initialized residual path gradually
incorporates observation-aligned local details without disrupting the
token-derived prediction at the start of training. The geometry decoder
is used only for supervision and is removed at inference.


\newsavebox{\foreactlatencypanel}
\newsavebox{\foreacttablepanel}
\newsavebox{\foreactablationpanel}
\newlength{\foreactlatencygap}
\newlength{\foreactablationgap}
\newlength{\foreactactivegap}
\makeatletter
\def\foreactpageextract#1#2#3#4#5{#2}
\newcommand{\foreactpage}[1]{%
  \expandafter\expandafter\expandafter\foreactpageextract\csname r@#1\endcsname}
\AddToHook{cmd/@startcolumn/before}{%
  \global\foreactactivegap=0pt
  \if@firstcolumn
    \@ifundefined{r@fig:real_world_eval}{}{%
      \ifnum\value{page}=\foreactpage{fig:real_world_eval}\relax
        \global\foreactactivegap=\foreactlatencygap
      \fi}%
    \@ifundefined{r@tab:libero_ablation}{}{%
      \ifnum\value{page}=\foreactpage{tab:libero_ablation}\relax
        \global\foreactactivegap=\foreactablationgap
      \fi}%
    \global\advance\@colht by \foreactactivegap
  \fi
}
\AddToHook{cmd/@makecol/after}{%
  \ifdim\foreactactivegap>0pt
    \setbox\@outputbox=\vbox{%
      \vskip-\foreactactivegap
      \box\@outputbox}%
    \global\advance\@colht by -\foreactactivegap
    \global\foreactactivegap=0pt
  \fi
}
\makeatother

\section{Experiments}
\label{sec:experiments}

In this section, we evaluate ForeAct on simulation and real-world manipulation tasks. We first describe the
experimental setup (Sec.~IV-A), then report simulation and real-world results (Secs.~IV-B and IV-C), analyze
inference latency (Sec.~IV-D), and study the contribution of the key components through ablations
(Sec.~IV-E).

\begin{figure*}[!t]
    \centering
    \includegraphics[width=\textwidth]
    {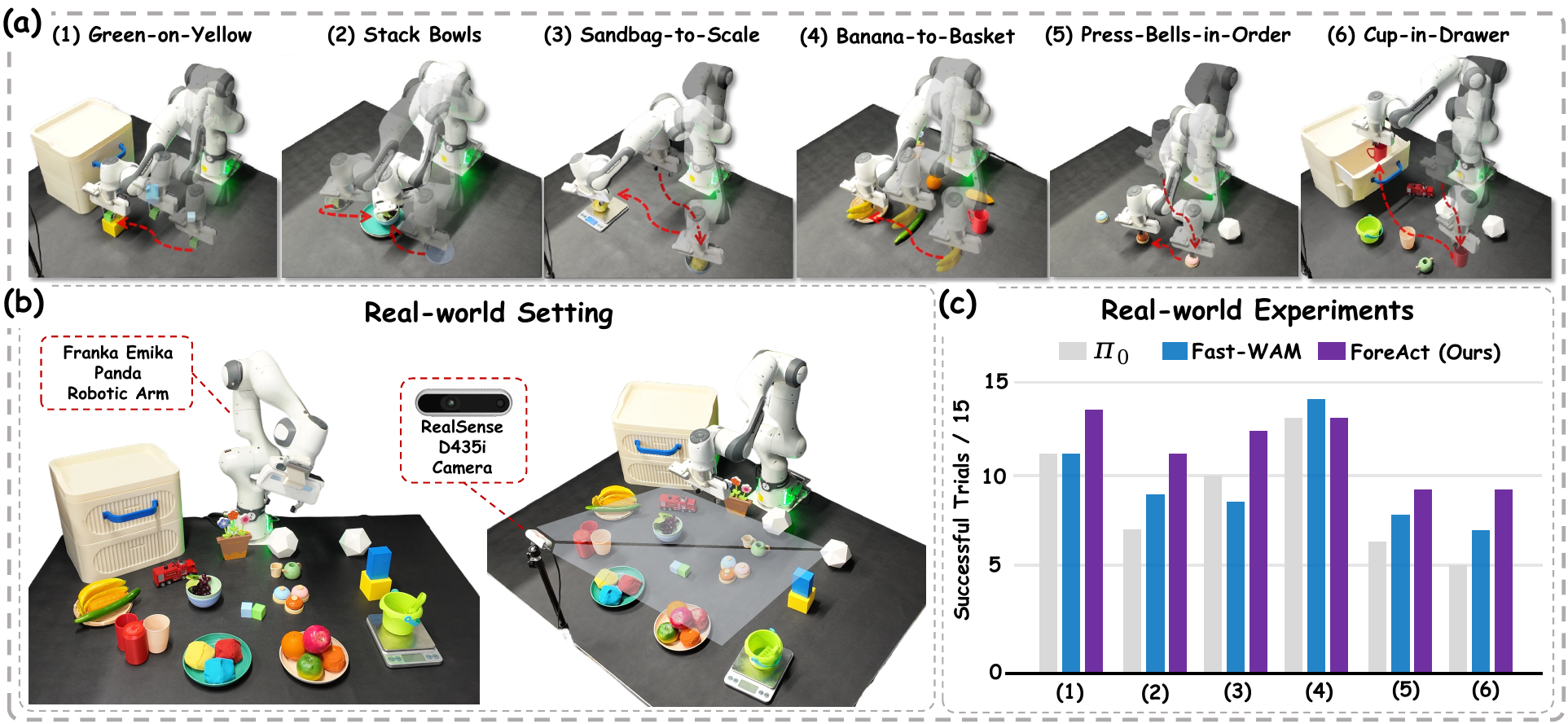}
    \caption{
        \textbf{Real-world evaluation.}
        (a) Six tabletop manipulation tasks spanning object placement,
        stacking, forceful interaction, and multi-stage execution:
        Green-on-Yellow, Stack Bowls, Sandbag-to-Scale,
        Banana-to-Basket, Press-Bells-in-Order, and Cup-in-Drawer.
        (b) Experimental setup with a Franka Emika Panda manipulator
        observed by a fixed Intel RealSense D435i RGB camera.
        (c) Task-wise numbers of successful executions for $\pi_0$,
        FastWAM, and ForeAct over 15 trials per task.
        ForeAct achieves the best result on five of the six tasks and
        succeeds in 67 of 90 trials (74.4\%), compared with 58 (64.4\%)
        for FastWAM and 52 (57.8\%) for $\pi_0$.
    }
    \label{fig:real_world_eval}

    \sbox{\foreactlatencypanel}{%
        \begin{minipage}[t]{0.49\textwidth}
            \vspace{0pt}
            \centering
            \includegraphics[width=\linewidth]{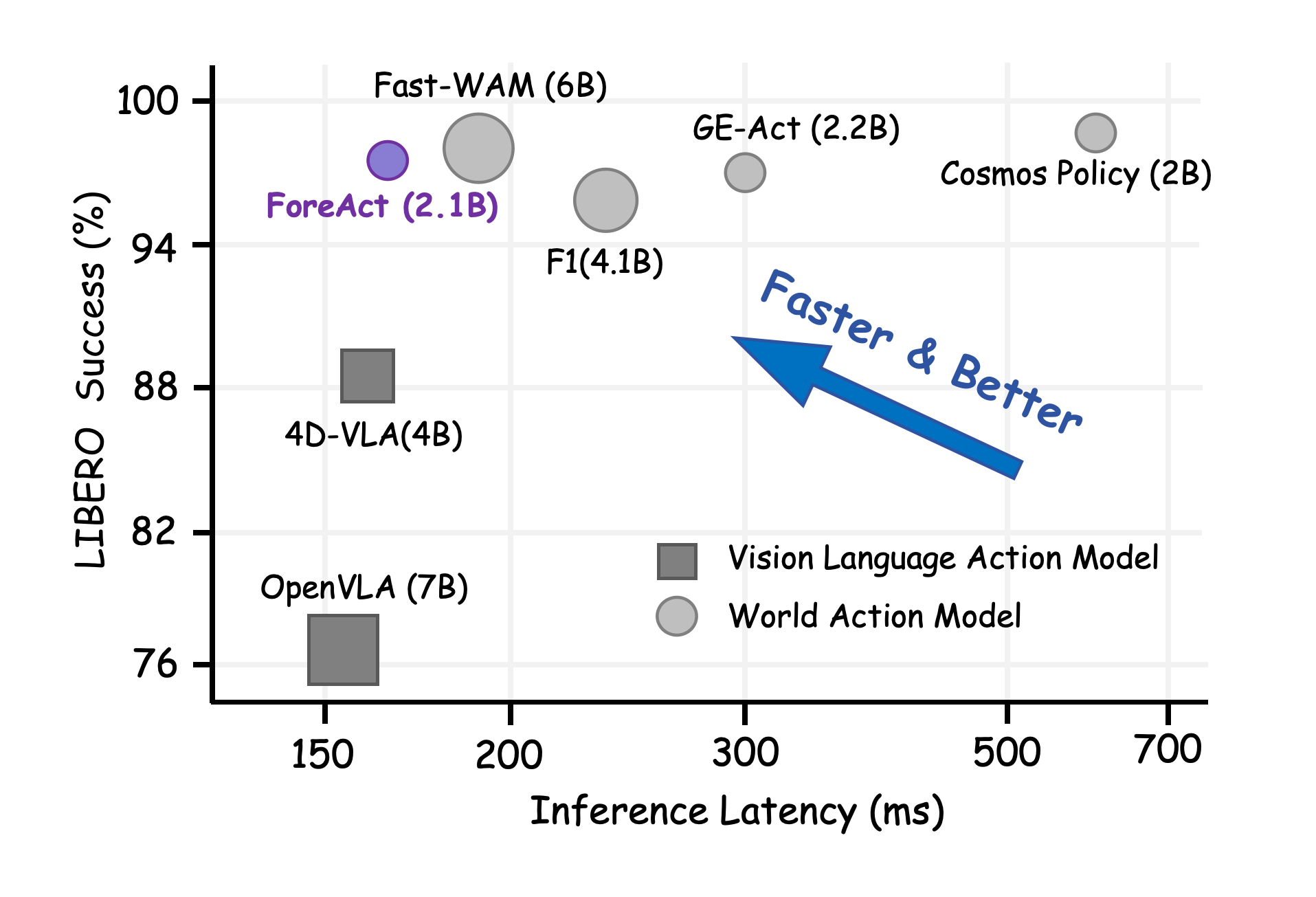}
            \captionof{figure}{
                \textbf{Success rate and inference latency on LIBERO.}
                Marker size indicates model size; circles denote WAMs
                and squares denote VLAs. Latency is shown on a
                logarithmic scale.
            }
            \label{fig:libero_efficiency}
        \end{minipage}%
    }

    \global\foreactlatencygap=\dimexpr
        \ht\foreactlatencypanel+
        \dp\foreactlatencypanel+
        \medskipamount
    \relax

    \par\medskip
    \noindent\hfill\usebox{\foreactlatencypanel}
\end{figure*}

\subsection{Experimental Setup}

\indent\textbf{Simulator Setup.}
We evaluate ForeAct on LIBERO~\cite{liu2023libero} and
SimplerEnv-Bridge~\cite{li2024simplerenv}. LIBERO uses a Franka
Emika Panda manipulator and comprises four suites that evaluate
generalization across spatial arrangements, objects, language-specified
goals, and long-horizon tasks. Each suite contains 500 demonstrations
across 10 tasks. SimplerEnv-Bridge evaluates visual-matching
manipulation with a WidowX robot under variations in scene appearance
and object configuration. We use fixed evaluation seeds of 42 for
LIBERO and 7 for SimplerEnv-Bridge. The complete evaluation protocol
is repeated three times for each benchmark, and performance is reported
as the mean success rate.

\begin{table}[t]
    \centering
    \renderliberocomparisontable
\end{table}

\begin{table}[t]
    \centering
    \rendersimplerenvcomparisontable
\end{table}

\newcommand{\renderliberoablationtable}{%
\begingroup
\centering

\captionof{table}{%
\textbf{Component and design ablations of ForeAct on LIBERO.}
\textit{WM State} distinguishes noisy diffusion latents from clean observation
latents. \textit{Geo. Tok.} and \textit{Attn.} specify the use and interaction
of learnable geometry tokens, while \textit{Aux. Sup.} denotes the auxiliary
prediction pathway. $\mathbf{V}$ and $\mathbf{Q}$ represent visual and geometry
tokens, respectively; arrows indicate information flow.
}
\label{tab:libero_ablation}

\setlength{\tabcolsep}{2.8pt}
\renewcommand{\arraystretch}{1.08}

\resizebox{\linewidth}{!}{
\begin{tabular}{lccccc}
\toprule
\textbf{Variant}
& \textbf{WM State}
& \textbf{Geo. Tok.}
& \textbf{Attn.}
& \textbf{Aux. Sup.}
& \textbf{Avg. $\uparrow$} \\
\midrule

Diffusion WAM
& Noisy & -- & -- & Future frames & 93.2 \\
Action Only
& Clean & \xmark & -- & -- & 94.3 \\
Geometry Tokens Only
& Clean & \cmark & $\mathbf{V}\leftrightarrow\mathbf{Q}$ & -- & 94.7 \\
Direct Feature Sup.
& Clean & \xmark & -- & $\mathbf{V}_t\!\to\!D_{t+1}$ & 93.4 \\
Masked Token Attn.
& Clean & \cmark & $\mathbf{Q}\leftarrow\mathbf{V}$ & $\mathbf{Q}_t\!\to\!D_{t+1}$ & 94.9 \\

\midrule
\rowcolor{ourscolor}
\textbf{ForeAct (Ours)}
& Clean & \cmark & $\mathbf{V}\leftrightarrow\mathbf{Q}$ & $\mathbf{Q}_t\!\to\!D_{t+1}$ & \textbf{96.6} \\

\bottomrule
\end{tabular}
}

\endgroup
}

\begin{figure*}[!t]
    \centering

    \sbox{\foreacttablepanel}{%
        \begin{minipage}[t]{0.49\textwidth}
            \vspace{0pt}
            \centering
            \renderliberoablationtable
        \end{minipage}%
    }

    \sbox{\foreactablationpanel}{%
        \begin{minipage}[t]{0.49\textwidth}
            \vspace{0pt}
            \centering
            \includegraphics[width=\linewidth]{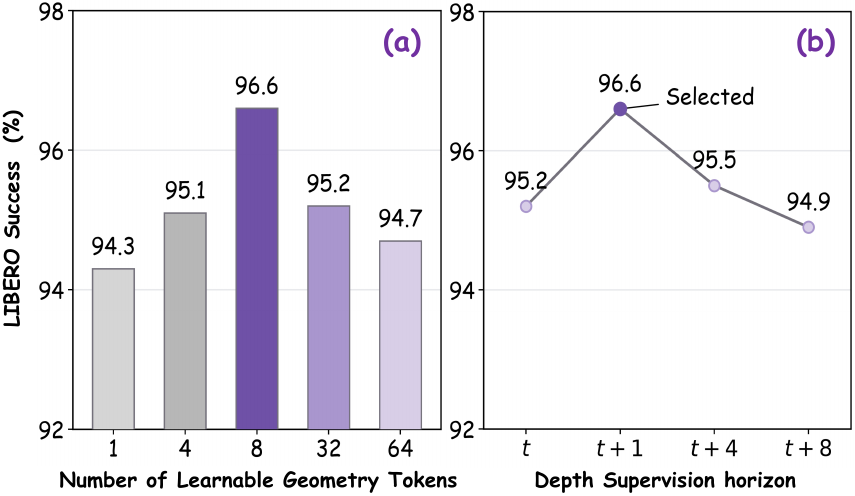}
            \captionof{figure}{
                \textbf{Sensitivity analysis of geometry-token count
                and depth-supervision horizon.}
                LIBERO average success under
                (a) different geometry-token counts and
                (b) different depth-supervision horizons.
            }
            \label{fig:ablation_sensitivity}
        \end{minipage}%
    }

    \global\foreactablationgap=\dimexpr
        \ht\foreactablationpanel+
        \dp\foreactablationpanel-
        \ht\foreacttablepanel-
        \dp\foreacttablepanel
    \relax

    \noindent
    \usebox{\foreacttablepanel}
    \hfill
    \usebox{\foreactablationpanel}
\end{figure*}

\indent\textbf{Real-World Setup.}
As illustrated in Fig.~\ref{fig:real_world_eval}, our physical
platform comprises a Franka Emika Panda arm and a fixed Intel
RealSense D435i RGB camera. We evaluate six tasks:
Green-on-Yellow and Banana-to-Basket test object transfer and
constrained placement; Stack Bowls tests precise alignment;
Sandbag-to-Scale involves contact-aware placement;
Press-Bells-in-Order requires ordered execution; and
Cup-in-Drawer combines grasping with articulated-object interaction.
All methods use the same monocular RGB view, action interface,
and execution protocol. We compare $\pi_0$, FastWAM, and ForeAct
using 15 rollouts per task (90 per method) and report successful
executions.

\indent\textbf{Implementation Details.}
ForeAct is initialized from the pretrained
Cosmos-Predict2-2B Video2World model~\cite{agarwal2025cosmos}.
We freeze the T5 text encoder and video VAE, while jointly training
the Cosmos video DiT, a 16-layer DiT-B action expert with hidden
dimension 1,024, and the geometric readout. Eight learnable global
geometry tokens are inserted into the Cosmos token stream and
supervised with MoGe-3~\cite{wang2025moge} normalized relative-depth
targets from the subsequent frame. MoGe-3 is used only for offline target generation and is not part
of the deployed policy. For LIBERO, we train for 100K
steps on four GPUs with a global batch size of 48; for
SimplerEnv-Bridge, we train for 50K steps on eight GPUs with a
global batch size of 96. The world-model backbone uses a learning
rate of $1\times10^{-5}$, while the action expert and geometric
readout use $1\times10^{-4}$. We set the action horizon to 8 for
LIBERO and 16 for SimplerEnv. The auxiliary geometric objective
is weighted by $0.01$.

\subsection{Evaluation on Simulator}
\label{sec:libero_eval}

We evaluate ForeAct on two established simulation benchmarks to assess generalization across task structures and robustness to visual and embodiment shifts.  Figure~\ref{fig:depth_visualization} visualizes simulator RGB observations and the model’s predicted depth for the next time step.  

\indent\textbf{LIBERO Benchmark Results.} LIBERO evaluates language-conditioned manipulation across four suites that vary spatial arrangement, object identity, goal semantics, and execution horizon. As reported in Table~\ref{tab:libero_comparison}, ForeAct obtains $98.4\%$, $98.8\%$, $97.2\%$, and $91.8\%$ on LIBERO-Spatial, LIBERO-Object, LIBERO-Goal, and LIBERO-Long, yielding a $96.6\%$ average success rate. It achieves the best Spatial score and the second-best Object and Goal scores, demonstrating strong transfer of action-relevant geometry. Among world action models without additional robot-action pre-training, ForeAct improves over WorldVLA and FlowVLA by $14.8\%$ and $8.5\%$, respectively, while remaining competitive with Fast-WAM and Cosmos Policy, with performance gaps of only $1.0\%$ and $1.9\%$. It also exceeds the geometry-aware DepthVLA and GeoPredict by $1.7\%$ and $0.1\%$ without additional depth input. The lower Long score reflects the accumulated difficulty of extended-horizon execution.

\indent\textbf{SimplerEnv Benchmark Results.}
SimplerEnv-Bridge complements LIBERO by evaluating visual-matching manipulation on the WidowX platform with BridgeData V2 demonstrations under variations in object appearance, placement, and visual context, testing whether policies preserve task-relevant geometry across such changes. As shown in Table~\ref{tab:simplerenv_comparison}, ForeAct ranks first overall with a $63.5\%$ average success rate, exceeding GR00T N1.5 by $1.6$ percentage points. It performs best on Spoon on Towel ($87.5\%$) and ties the best result on Eggplant in Basket ($100.0\%$), demonstrating robust spatial precision across these variations. ForeAct achieves $54.2\%$ on Carrot on Plate, whereas its lower performance on Stack Blocks ($12.5\%$) reflects the difficulty of sequential multi-object alignment. Overall, these results indicate that future-geometry supervision transfers useful spatial structure across visual variations without requiring depth input or an external geometry model at deployment.

\label{sec:simplerenv_eval}

\subsection{Evaluation on Real-World}
\label{sec:real_world_eval}

\textcolor{black}{We evaluate ForeAct on six tabletop tasks using a Franka Emika Panda arm, covering placement, transfer, forceful interaction, and multi-stage sequencing. Figure~\ref{fig:real_world_eval} compares ForeAct with $\pi_0$~\cite{black2024pi0} and FastWAM~\cite{yuan2026fastwam} under identical camera and execution settings to assess real-world manipulation performance.}

\textcolor{black}{\textbf{Real-World Performance.}} \textcolor{black}{As shown in Fig.~\ref{fig:real_world_eval}, ForeAct achieves the strongest aggregate success and leads on most tasks, with particularly clear gains on longer-horizon sequences requiring reliable object localization and geometric consistency across multiple actions. Its competitive performance on simpler single-stage tasks indicates that predictive geometric representations preserve basic execution accuracy. The consistent gains over $\pi_0$ and FastWAM further support shaping action features through future-geometry supervision within the shared video backbone.}


\subsection{Latency Analysis}
\label{sec:latency_analysis}

To assess inference efficiency, we compare ForeAct with representative VLAs and WAMs on LIBERO in terms of average success rate, latency, and parameter count, as shown in Fig.~\ref{fig:libero_efficiency}. Latency is evaluated following the corresponding inference settings on a single NVIDIA H20 GPU with batch size 1, covering observation encoding through action-chunk generation.

\textcolor{black}{\textbf{Inference Efficiency.}}
\textcolor{black}{ForeAct achieves a $96.6\%$ average success rate at $168$\,ms with approximately $2.1$B parameters, placing it in a favorable accuracy--latency regime. It approaches the accuracy of larger world action models with fewer parameters and lower latency, while substantially outperforming the faster conventional VLA baseline. At a comparable success rate, its latency is roughly one-third that of Cosmos Policy, and it is also faster than Fast-WAM while using a considerably smaller model. This efficiency comes from extracting predictive representations and decoding actions with a single world-model backbone pass, without iterative future-frame denoising, making ForeAct suitable for closed-loop manipulation.}

\subsection{Ablation Study}

\textcolor{black}{
We ablate ForeAct on LIBERO to evaluate single-pass world
representations and token-mediated future-geometry supervision.
Table~\ref{tab:libero_ablation} examines component contributions,
while Fig.~\ref{fig:ablation_sensitivity} analyzes geometry-token
capacity and supervision horizon.
}

\textcolor{black}{\textbf{Components and Design.}}
\textcolor{black}{
Table~\ref{tab:libero_ablation} shows that the clean action-only
baseline outperforms Diffusion WAM ($94.3\%$ vs.\ $93.2\%$),
supporting a single video-backbone pass for action prediction
without iterative video denoising. Adding geometry tokens without
future-depth supervision yields a modest gain to $94.7\%$, whereas
incorporating future-depth supervision raises success to $96.6\%$.
This result indicates that the improvement extends beyond additional
token capacity. The supervision pathway is also important. Directly
imposing depth supervision on the visual features yields only
$93.4\%$, supporting a dedicated geometry-token interface.
Restricting attention to one-way visual-to-geometry information flow
reduces success to $94.9\%$, supporting bidirectional interaction
through which depth-supervised geometry tokens can shape the visual
features before action decoding. Together, these results demonstrate
the complementary benefits of clean world representations and
token-mediated geometric supervision.
}

\textcolor{black}{\textbf{Geometry Tokens and Supervision Horizon.}}
\textcolor{black}{
Fig.~\ref{fig:ablation_sensitivity}(a) shows that increasing the
geometry-token count from one to eight improves success from
$94.3\%$ to $96.6\%$, whereas expanding it to 32 and 64 lowers
performance to $95.2\%$ and $94.7\%$, respectively. This suggests
that a compact token set provides sufficient capacity for effective
geometric supervision, while additional tokens provide no further
gain. Fig.~\ref{fig:ablation_sensitivity}(b) shows that next-frame
depth supervision outperforms current-frame depth supervision
($96.6\%$ vs.\ $95.2\%$), supporting the value of anticipating
spatial changes beyond encoding observed geometry. Extending the
supervision horizon to $t+4$ and $t+8$ reduces success to $95.5\%$
and $94.9\%$, suggesting that immediate-future geometry provides a
more effective learning signal under the single-observation
formulation. These findings support the default choice of eight
geometry tokens with next-frame depth supervision.
}
\section{Conclusion}

\begingroup
\color{black}


We presented ForeAct, a unified backbone combining pretrained spatiotemporal priors with future-geometry supervision for efficient robotic manipulation. ForeAct extracts action features from clean observation latents in one video-backbone pass, without future-frame synthesis or iterative denoising. Learnable geometry tokens receive next-frame relative-depth supervision and interact with visual features through bidirectional self-attention within the shared video backbone. Depth is used only for training, so deployment requires neither depth observations nor a separate geometric backbone. Experiments on LIBERO, SimplerEnv-Bridge, and real-world tasks demonstrate strong performance in simulation and physical settings. Ablations support the complementary roles of the representation and geometry-supervision designs. Together, these results show that pretrained video world models can serve as geometry-aware backbones for efficient, low-latency robot control.

\endgroup

\bibliographystyle{ieeetr}
\bibliography{references/references}

\end{document}